\documentclass{article}
\usepackage{multirow}
\usepackage{spconf,amsmath,graphicx}
\usepackage{url}

\usepackage{amssymb}
\usepackage{multirow, threeparttable, booktabs, makecell}
\usepackage{color}
\usepackage{graphicx}
\usepackage{enumitem}
\usepackage{array}
\usepackage{placeins}
\usepackage{afterpage}
\usepackage[numbers,sort&compress]{natbib}

\title{MF-AED-AEC: SPEECH EMOTION RECOGNITION BY LEVERAGING MULTIMODAL FUSION, ASR ERROR DETECTION, AND ASR ERROR CORRECTION}
\name{Jiajun He$^{\star}$\footnotemark, Xiaohan Shi$^{\star}$\footnotemark, Xingfeng Li$^{\dagger}$, Tomoki Toda$^{\ddagger}$}
\address{$^{\star}$ Graduate School of Informatics, Nagoya University \\ $^{\dagger}$ School of Computer Science and Technology, Hainan University \\ $^{ \ddagger}$ Information Technology Center, Nagoya University}

\begin{document}
\ninept
\maketitle
\begin{abstract}
The prevalent approach in speech emotion recognition (SER) involves integrating both audio and textual information to comprehensively identify the speaker's emotion, with the text generally obtained through automatic speech recognition (ASR).
An essential issue of this approach is that ASR errors from the text modality can worsen the performance of SER. Previous studies have proposed using an auxiliary ASR error detection task to adaptively assign weights of each word in ASR hypotheses. However, this approach has limited improvement potential because it does not address the coherence of semantic information in the text.
Additionally, the inherent heterogeneity of different modalities leads to distribution gaps between their representations, making their fusion challenging.
Therefore, in this paper, we incorporate two auxiliary tasks, ASR error detection (AED) and ASR error correction (AEC), to enhance the semantic coherence of ASR text, and further introduce a novel multi-modal fusion (MF) method to learn shared representations across modalities.
We refer to our method as MF-AED-AEC. Experimental results indicate that MF-AED-AEC significantly outperforms the baseline model by a margin
of 4.1\%.


\makeatletter
\let\@makefnmark\relax
\makeatother
\footnotetext{These authors annotated with $^{\star}$ contributed equally to this work.}

\end{abstract}
\section{Introduction}
\label{sec:intro}
Multimodal speech emotion recognition (SER) aims to identify and comprehend human emotions by integrating information from various perceptual modalities, such as audio, text, video, and images. SER 
has gained extensive attention due to its wide-ranging applications in fields like human-computer interaction, healthcare, and intelligent customer service \cite{ al2023speech, de2023ongoing}.

In this paper, we focus on the two most common modalities (speech and text) for multimodal SER.
For the speech modality, early research focused on extracting low-level features such as Mel-frequency cepstral coefficients (MFCCs) and filter banks (FBanks), or handcrafted features \cite{ghosh2022mmer}. 
Recently, with the advancement of deep learning, recurrent neural networks (RNNs) \cite{liu2022group}, convolutional neural networks (CNNs) \cite{fan2022isnet}, and transformer models \cite{chen2022speechformer} have significantly improved the performance of SER \cite{fan2023mgat}. 
Furthermore, with the notable success of self-supervised learning (SSL), speech-based pretrained models like wav2vec \cite{schneider2019wav2vec}, HuBERT \cite{hsu2021hubert}, and WavLM \cite{chen2022wavlm} have achieved state-of-the-art (SOTA) performance in SER.

For the text modality, while text-based pretrained models such as BERT \cite{devlin2018bert}, DeBERTa \cite{he2020deberta}, and RoBERTa \cite{liu1907roberta} have demonstrated excellent performance in multimodal SER tasks, the accuracy of automatic speech recognition (ASR) results is equally crucial for achieving accurate SER result \cite{lin2023robust, shi2023effectiveness}. To mitigate the impact of ASR errors, Santoso et al. combined self-attention mechanisms with word-level confidence measures to reduce the importance of high-error probability words, thus improving SER performance \cite{santoso2022speech}. 
However, this approach is heavily reliant on the performance of the ASR system and lacks generalization ability.
Additionally, Lin et al. introduced an auxiliary ASR error detection task to determine the probabilities of erroneous words, indicating how much trust should be placed in each word of ASR hypotheses, thereby enhancing multimodal SER performance \cite{lin2023robust}. However, this method solely focuses on error detection in ASR hypotheses without correction, which means it does not improve the coherence of semantic information. Therefore, it has certain limitations in enhancing SER results.

On the other hand, for multimodal SER tasks, how to perform multimodal fusion between the speech and text modalities is another key focus. Previous approaches have included simple feature concatenation \cite{tripathi2018multi}, CNNs \cite{choi2018convolutional}, and cross-modal attention \cite{krishna2020multimodal}, among others. While these methods have shown some effectiveness, they often face challenges arising from the representation gap between different modalities.
In recent multimodal tasks like speech recognition \cite{hu2023mir} and sentiment analysis \cite{hazarika2020misa, yu2021learning, yao2022modality}, researchers have proposed learning two distinct representations to enhance multimodal learning. The first representation is modality-invariant, mapping all modalities of the same utterance into a shared subspace with distributional alignment. This captures the commonalities of multimodal signals and the shared motives and goals of the speaker, which influence the emotional state of the utterance. Additionally, they learn modality-specific representations tailored to each modality. Combining these two representations provides a comprehensive perspective on multimodal data for downstream tasks \cite{yang2022learning}.

Motivated by these aforementioned observations, we propose a multi-task learning approach for multimodal SER based on two auxiliary tasks: ASR error detection (AED) and ASR error correction (AEC). Specifically, we introduce an AED module to identify the locations of ASR errors. Subsequently, we employ an AEC module to correct these errors, thereby reducing the impact of ASR errors on SER tasks.
In addition, we also design a multimodal fusion module to learn modality-specific and modality-invariant representations within a shared audio-textual modality space. These two representations are then fused and utilized for downstream SER tasks. Empirical results demonstrate the effectiveness of this approach. In summary, our main contributions are as follows:
\begin{figure*}[htbp]
  \centering
  \includegraphics[scale=0.143]{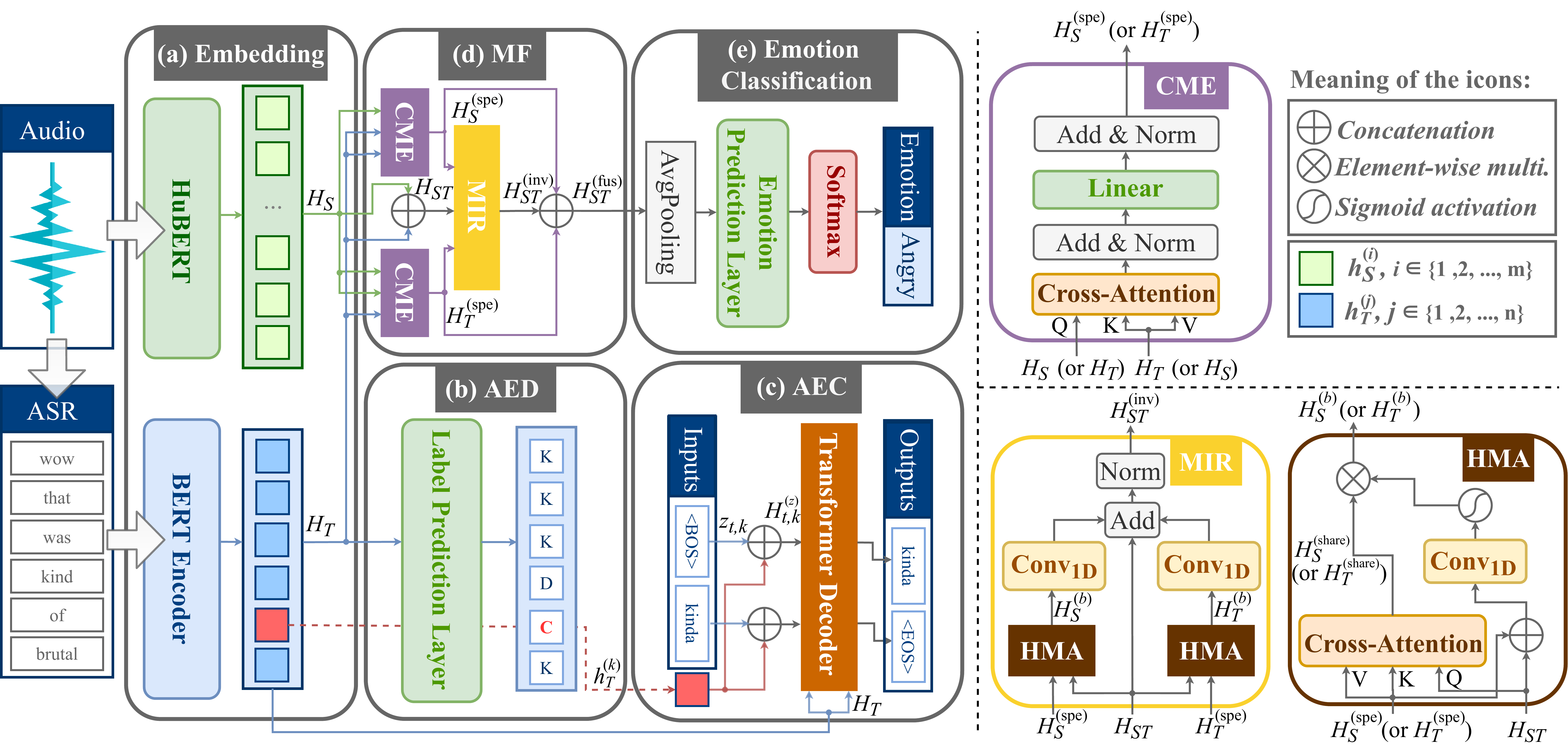}
  
  \caption{Overall architecture of the proposed MF-AED-AEC model.}
  \label{fig:model}
\end{figure*}

\begin{itemize}[leftmargin=*]
\vspace{-2mm}
\setlength{\topsep}{0pt}
\setlength{\itemsep}{0pt}
\setlength{\parsep}{0pt}
\setlength{\parskip}{0pt}
\item We introduce a multi-task learning method by incorporating two auxiliary tasks focused on ASR error detection and correction. This approach improves the coherence of semantic information of ASR hypotheses, thereby augmenting SER performance.
\item We propose a multi-modal fusion approach that leverages shared modality characteristics to bridge the gap between heterogeneous modalities, effectively enhancing SER performance.
\item Our proposed MF-AED-AEC method outperforms previous baseline models by a large margin on the IEMOCAP dataset.
\end{itemize}

\vspace{-2mm}
\section{Proposed Method}
\label{sec:format}


\subsection{Problem Formulation}
\label{section2.1}
The multimodal multi-task SER challenge we address can be expressed as the function $f(S, T) = (L, C)$, where the speech modality $S = (s_1, s_2, \cdots , s_m)$ consists of \textit{m} frames extracted from an utterance. The text modality $T = (t_1, t_2, \cdots , t_n)$ represents the original ASR hypotheses of an utterance, comprising $n$ tokens. All tokens are mapped to a predefined WordPiece vocabulary \cite{wu2016google}. Additionally, operating within our multitask learning framework, the primary task focuses on emotion classification, yielding output $L \in \{l_1, l_2, \cdots , l_e\}$, where $e$ represents the emotional categories. Concurrently, the auxiliary tasks encompass AED and AEC. The outcome of these auxiliary tasks is $C = (c_1, c_2, \cdots , c_p)$, denoting the human-annotated transcripts of the utterance, comprising $p$ tokens.

\vspace{-2mm}
\subsection{Embedding Module}
\label{section2.2}
Our embedding module consists of acoustic embedding and token embedding, as illustrated in Fig. \ref{fig:model}. In this section, we provide a detailed explanation.

\textbf{Contextual Speech Representations.} To acquire comprehensive contextual representations of acoustic features, we leverage a pretrained SSL model, HuBERT \cite{hsu2021hubert} as our acoustic encoder. HuBERT employs a combination of CNN layers and a transformer encoder to capture both speech features and contextual context. 
We adopt $H_S = (h_S^{(1)}, h_S^{(2)}, \cdots , h_S^{(m)})$ to symbolize the acoustic hidden representations generated through HuBERT.


\textbf{Contextual Token Representations.} The hidden representations  $H_T = (h_T^{(1)}, h_T^{(2)}, \cdots , h_T^{(n)})$ of the model inputs $T$ are obtained by using the pretrained language model BERT \cite{devlin2018bert} as our text encoder.

\begin{equation}
    H_T = {\rm BERT}({\rm TE}(T)+{\rm PE}(T)),
\label{eq3}
\end{equation}
where TE and PE denote the token embedding and position embedding, respectively. 

\vspace{-2mm}
\subsection{ASR Error Detection (AED) Module}

Similarly to \cite{DBLP:journals/corr/abs-2208-04641}, we align $T$ and $C$ by determining the longest common subsequence (LCS) between them. The aligned tokens are labeled \textit{KEEP} (\textbf{K}), whereas the remaining tokens are labeled \textit{DELETE} (\textbf{D}) or \textit{CHANGE} (\textbf{C}).
The label prediction layer is a straightforward fully connected (FC) network with three classes.
\begin{equation}
\label{eq2}
  P(y_o|h_T^{(o)}) = {\rm SoftMax}({\rm FC}(h_T^{(o)})),
\end{equation}
where $h_T^{(o)} \in H_T$ and $y_o$, $o \in \{1, \cdots, n\}$ are the output of the BERT encoder and predicted labeling operations, respectively.

\subsection{ASR Error Correction (AEC) Module}

Our decoder operates in parallel to the tokens predicted as \textbf{C}.
For the $k^{th}$ change position, the decoding sequence can be represented as $Z_k = (z_{1,k}, z_{2,k}, \cdots , z_{d,k})$, where $d$ is the length of the decoding sequence, generated by the transformer decoder. We compute the decoder inputs at step $t$ as follows:
\begin{equation}
  H_{t,k}^{(z)} = {\rm FC}(({\rm TE}(z_{t,k})+{\rm PE}(z_{t,k})) \oplus h_T^{(k)}),
\end{equation}
where TE and PE are the same token embedding and position embedding as in Eq. (\ref{eq3}), respectively. $z_{1,k}$ is initialized by a special start token $<$\textit{BOS}$>$. $h^{(k)}_T$ is the output of the BERT encoder at the $k^{th}$ change position. ``$\oplus$" denotes a concatenate function. 
Then, a generic transformer decoder is applied to obtain the decoder layer output, where the query input $Q$ is the decoder input. Both the key input $K$ and the value input $V$ are the hidden representations of the BERT encoder \cite{vaswani2017attention}:

\begin{equation}
%
O_{t+1,k}^{{\rm(gen)}} = {\rm Transformer_{Decoder}}(H_{t,k}^{(z)},H_T,H_T),
\end{equation}
where $O_{t+1,k}^{\rm(gen)}$ is the decoder layer output. 
Finally, the generation output is calculated as:

\begin{equation}
P_{t+1,k}^{\rm(gen)} = {\rm SoftMax}({\rm FC}(O_{t+1,k}^{\rm(gen)})).
\end{equation}

\subsection{Multimodal Fusion (MF) Module}
Inspired by \cite{hu2023mir}, our multimodal fusion (MF) module is composed of two cross-modal encoder (CME) blocks and one modality-invariant representations (MIR) block. The objective is to facilitate the learning of modality-specific representations and modality-invariant representations.
In this section, we provide an in-depth explanation of the operation of each CME block and the MIR block.

\textbf{CME Block} is structured akin to a standard transformer layer, featuring an \textit{h}-head cross-attention module \cite{tsai2019multimodal}, residual connections, and FC layers.
In order to acquire speech-aware token representations and token-aware speech representations, we utilize two CME blocks. This is achieved by employing $H_T$ ({\rm or} $H_S$) as queries and $H_S$ ({\rm or}  $H_T$) as keys and values within each CME block.

\vspace{-3mm}
 \begin{equation}
Q = H_S \, ({\rm or} \, H_{T}) ,K=H_T \, ({\rm or} \, H_S),V=H_T \, ({\rm or} \, H_S)
\end{equation}
%
\begin{equation}
H_S^{{\rm(spe)}} \, ({\rm or} \, H_T^{\rm(spe)}) = {\rm FC}({\rm Cross \text{-}}{\rm Attention}(Q,  K, V)).
\end{equation}

\textbf{MIR Block} utilizes a hybrid-modal attention (HMA) module to extract the shared information from each modality-specific representation that pertains to both modalities:
\begin{equation}
H_i^{(b)} = {\rm HMA}( H_i^{\rm(spe)}, H_{ST}), i \in \{S,T\},
\end{equation}
where $i$ denotes either the speech or text modality, and $H_{ST}$ signifies the concatenation of $H_{S}$ and $H_{T}$, comprising both speech and text information. The resulting features are subsequently summed with $H_{ST}$, culminating in the ultimate modality-invariant representation:
\begin{equation}
H_{ST}^{\rm(inv)} = {\rm Norm}( H_{ST} + \sum_{i \in \{S,T\}} {\rm Conv_{1d}}(H_i^{(b)})),
\end{equation}
where the “{\rm Norm}” represents layer normalization \cite{ba2016layer}, “${\rm Conv_{1d}}$” denotes $1 {\rm \times} 1$ convolution followed by PReLU activation \cite{he2015delving}.

\textbf{HMA Block} initiates with a cross-attention layer aimed at extracting the shared information from each modality-specific representation, pertinent to both modalities:
\begin{equation}
H_i^{\rm(share)} = {\rm Cross \text{-}}{\rm Attention}(H_{ST},  H_i^{\rm(spe)}, H_i^{\rm(spe)}), i \in \{S,T\}.
\end{equation}

To enhance the feature's modality invariance, a parallel convolutional network is employed to learn a mask that filters out modality-specific information:
\begin{equation}
H_i^{(b)} = H_i^{\rm(share)} \otimes \sigma ({{\rm Conv_{1d}}(H_i^{\rm(spe)} \oplus H_{ST})}), i \in \{S,T\},
\end{equation}
where “$\sigma$” denotes Sigmoid activation and “$\otimes$” indicates element-wise multiplication.
Finally, the modality-specific and modality-invariant representations are concatenated together to get the final multimodal fusion representations $H_{ST}^{\rm(fus)}$:

\begin{equation}
H_{ST}^{\rm(fus)} = H_S^{\rm(spe)} \oplus H_T^{\rm(spe)} \oplus H_{ST}^{\rm(inv)}.
\end{equation}

\subsection{Emotion Classification Module}
The emotion classification is performed by applying the temporal average pooling layer on the output feature $H_{ST}^{\rm (fus)}$ of the MF module, followed by an FC layer and a SoftMax activation function.

\vspace{-3mm}
\begin{equation}
\label{eq13}
  P(y_{\rm emo}|H_{ST}^{\rm(fus)}) = {\rm SoftMax}({\rm FC}({\rm AvgPooling}(H_{ST}^{\rm(fus)}))),
\end{equation}
where $y_{\rm emo}$ is the predicted emotion classification.

\subsection{Joint Training}

The learning process is optimized through three cross entropy loss functions that correspond to emotion classification, AED, and AEC.
\begin{equation}
{\rm Loss}_{\rm emo} = - \sum{\rm log}(P(y_{\rm emo}|H_{ST}^{\rm(fus)}))
\end{equation}
\vspace{-3mm}
\begin{equation}
{\rm Loss}_d = - \sum_{o}{\rm log}(P(y_o|h_T^{(o)}))
\end{equation}
\begin{equation}
\begin{split}
{\rm Loss}_e = - \sum_{k}\sum_{t}{\rm log}(P_{t,k}^{\rm(gen)}) .
\end{split}
\end{equation}

These three loss functions are linearly combined as the overall objective during the training stage:
\begin{equation}
{\rm Loss} = {\rm Loss}_{\rm emo} + \beta \cdot (\gamma \cdot {\rm Loss}_d + {\rm Loss}_e),
\end{equation}
where $\gamma$ is the hyperparameter for adjusting the weight between ${\rm Loss}_d$ and ${\rm Loss}_e$ and $\beta$ is the hyperparameter for adjusting the weight between main task and auxiliary tasks.

During the inference stage, the AED and AEC modules are excluded. The remaining network accepts speech data and ASR hypotheses as input and outputs emotion classification results.

\section{Experiments and Results}
\label{sec:majhead}

\subsection{Experiment Settings}
\label{ssec:Experiment Settings}
Our method was implemented using Python 3.10.0 and Pytorch 1.11.0. The model was trained and evaluated on a computer with Intel(R) Xeon(R) Gold 6248 CPU @ 2.50GHz, 32GB RAM and one NVIDIA Tesla V100 GPU. 
The acoustic encoder was initialized using hubert-base-ls960, ultimately yielding acoustic representations characterized by a dimensionality of 768. 
The text encoder employed bert-base-uncased model for initialization. The vocabulary size for word tokenization was set to 30522. We set the hidden size as 768, the number of attention layers as 12, and the number of attention heads as 12. Both the HuBERT model and the BERT model were finetuned during the training stage.
The transformer decoder adopted a single-layer transformer with a hidden size of 768.
We used Adam \cite{kingma2014adam} as the optimizer with a batch size of 16. For training, we kept the learning rate constant at $1e^{-5}$, which worked well for all our configurations. For our multi-task learning setup, we set $\gamma$ to 3 and set $\beta$ to 0.1.
To evaluate the classification performance, we used unweighted average recall (UAR).

\vspace{-2mm}
\subsection{Dataset}
\vspace{-1mm}
\label{ssec:Dataset}
To evaluate the effectiveness of our proposed model, we carried out experiments on the Interactive Emotional Dyadic Motion Capture (IEMOCAP) dataset \cite{busso2008iemocap} \footnote{\url{https://sail.usc.edu/iemocap/}}, which comprised roughly 12 hours of speech from ten speakers participating in five scripted sessions. 
Consistent with prior work in SER, we employed 5531 utterances that were categorized into four emotion categories: ``neutral" (1708), ``angry" (1103), ``happy" (including ``excited") (595 + 1041), and ``sad" (1084). 
Likewise, we conducted experiments with the 5-fold leave-one-session-out cross-validation.


\vspace{-2mm}
\subsection{Results and Analysis}
\vspace{-1mm}
\label{ssec:Results and Analysis}

Our experimental results are shown in Table \ref{tab:results}, where we conduct experiments on both single-modal and multi-modal setups. For the single-modal experiments, we select the HuBERT model and the BERT model as speech and text baseline models, respectively. The text includes human-annotated transcripts and ASR hypotheses obtained from the ``openai/whisper-medium.en" in the Whisper ASR model \cite{radford2022robust} with a word error rate (WER) of 20.48\% on the IEMOCAP dataset. All baseline models employ the same emotion classification module, as depicted in Fig. \ref{fig:model} (e). 
Our proposed single-modal model operates on the text modality, achieving emotion classification through multi-task joint training involving the emotion classification task, ASR error detection task, and ASR error correction task.
It is evident that while our proposed model's UAR is lower than that of the single-speech modality, the disparity when compared to using transcripts is merely 0.1\%. Moreover, our proposed model's UAR is 1.2\% higher than the UAR from single-task learning.

\vspace{-2mm}
\begin{table}[h!]
  \caption{Comparison results to baseline networks on IEMOCAP. 
  }
    \label{tab:results}
  \centering
  \resizebox{\linewidth}{!}{
  \begin{tabular}{@{}c|ccc@{}}
    \toprule
    \multirow{1}{*}{\textbf{Method}}  & \textbf{Model} & \textbf{Modality} & \textbf{UAR (\%)}    \\
    \midrule
    \multirow{5}{*}{Single-modal}  &HuBERT &Speech  &69.8   \\
    \addlinespace[2pt] 
    \cline{2-4}
    \addlinespace[2pt] 
     & BERT &Text (Transcripts)  &67.7 \\
     &BERT &Text (ASR)  &66.4 \\
     &\textbf{BERT + AED}  &\multirow{2}{*}{\textbf{Text (ASR)}}  &\multirow{2}{*}{\textbf{67.6}} \\
     & \textbf{+ AEC (Proposed)} & &\\
   \midrule
   \multirow{3}{*}{Multi-modal}      &HuBERT + BERT &Speech + Text (Transcripts)    &75.5\\
   &HuBERT + BERT &Speech + Text (ASR)    &75.2\\
   
    &\textbf{MF-AED-AEC (Proposed)} &\textbf{Speech + Text (ASR)}   &\textbf{79.3}\\

    \bottomrule
  \end{tabular}
  }

\end{table}

Furthermore, in the multi-modal setup, we choose the baseline model that combines the aforementioned acoustic and text representations. We applied average pooling operations separately to the acoustic and text representations and concatenate them, followed by an FC layer and SoftMax activation function to compute emotion classification results. Evidently, the multi-modal baseline model demonstrates superior performance in contrast to the single-modal models, thereby highlighting the benefits derived from utilizing multi-modal inputs. Our proposed MF-AED-AEC achieves the best performance, even surpassing the UAR value achieved using human-annotated transcripts (75.5\%) by a large margin of 3.8\%, demonstrating the marked effectiveness of our model. 

The differences between our proposed MF-AEC-AED model and the baseline model are mainly reflected in three modules: the AED module, the AEC module, and the MF module. 
To determine the impact of these modules on emotion recognition task performance, we conduct a series of ablation experiments, and the results of these ablation studies are presented in Table \ref{tab:Ablation_study}.

Firstly, we discuss the impact of the AED module. Specifically, we remove the AED module and only introduce the AEC module as an auxiliary task. This requires the model to correct errors in each utterance from scratch, rather than solely correcting errors detected. It is noticeable that in both single-modal and multi-modal models, the absence of the AED module leads to a notable decrease in UAR results. The UAR of the single-modal model decreases by 1.3\%, while the UAR of the multi-modal model experiences a decline of 1.5\%. This demonstrates the necessity of the AED module. 
Additionally, we note that the UAR result (66.3\%) with only the AEC module in the single-modal case is even worse than the baseline result without this module (66.4\%). This phenomenon occurs because directly using a neural machine translation (NMT) model for AEC can even increase the WER \cite{leng2021fastcorrect}. Unlike NMT, which often requires modification of almost all input tokens, AEC involves fewer modifications but is more challenging. 
Hence, we need to consider the characteristics of ASR output and thoughtfully design models for AEC, which is why we introduced the two auxiliary tasks of AED and AEC.

\vspace{-3mm}
\begin{table}[h!]
    \caption{Impacts of different modules on the proposed model.}
  \label{tab:Ablation_study}
  \centering
  \resizebox{\linewidth}{!}{
  \begin{tabular}{@{}c|ccc@{}}
    \toprule
    \multirow{1}{*}{\textbf{Method}}  & \textbf{Model} & \textbf{Modality} & \textbf{UAR (\%)}    \\
    \midrule
    \multirow{3}{*}{Single-modal}  &\textbf{Proposed} &\textbf{Text (ASR)}  &\textbf{67.6}   \\
     & w/o AED &Text (ASR)  &66.3 \\
     & w/o AEC &Text (ASR)  &66.6 \\

   \midrule
   \multirow{4}{*}{Multi-modal} &\textbf{Proposed} &\textbf{Speech + Text (ASR)}    &\textbf{79.3}\\
   &w/o AED &Speech + Text (ASR)    &77.8\\
   &w/o AEC &Speech + Text (ASR)    &78.2\\
   &w/o MF &Speech + Text (ASR)    &77.4\\

    \bottomrule
  \end{tabular}
  }

\end{table}


Secondly, we discuss the impact of the AEC module. Similarly, for both single-modal and multi-modal models, without the AEC module, there is a significant decrease in UAR results. The single-modal model's UAR drops by 1.0\%, and the multi-modal model's UAR drops by 1.1\%, demonstrating the necessity of the AEC module. This module improves emotion recognition performance by correcting erroneous text positions.

Finally, we discuss the impact of the MF module. Specifically, in the absence of the MF module, we apply average pooling operations separately to acoustic and text representations, followed by a concatenation operation. It can be seen that without the MF module, UAR performance decreases by 1.9\%, demonstrating the effectiveness of our multi-modal fusion module. Interestingly, even without MF module, the UAR result (77.4\%) is higher than that obtained with transcripts (75.5\%). The higher performance may be attributed to the AED and AEC modules enhancing generalization, particularly valuable in the multi-modal scenario. These modules enable the model to better handle variations and nuances present in real transcriptions compared to the single modality setup.

In conclusion, through ablation experiments, we confirm the contributions of the AED, AEC, and MF modules in enhancing the performance of emotion recognition tasks. The introduction and integration of these modules enable our MF-AEC-AED model to achieve superior performance in multi-modal SER tasks, showcasing its flexibility and robustness, especially in the presence of ASR errors.

\vspace{-2mm}
\section{CONCLUSION}
\label{sec:print}
In this paper, we propose MF-AEC-AED, a novel multimodal multi-task SER method. MF-AEC-AED leverages a novel multimodal fusion network to learn modality-specific representations and modality-invariant representations. In addition, we design two auxiliary tasks, namely AED and AEC, aimed at enhancing the coherence of semantic information within the text modality. Results on the IEMOCAP dataset validate the effectiveness of the proposed method to ASR errors. In the future, we plan to introduce auxiliary tasks of visual modality and contrastive learning to further improve the performance of emotion recognition.

\noindent \textbf{Acknowledgements.} This work was supported in part by JST CREST Grant Number JPMJCR22D1, Japan, and JSPS KAKENHI Grant Number 21H05054.

\bibliographystyle{IEEEbib}
\bibliography{strings,refs}

\begin{thebibliography}{10}

\bibitem{al2023speech}
M.~J. Al-Dujaili and A.~Ebrahimi-Moghadam,
\newblock ``Speech emotion recognition: a comprehensive survey,''
\newblock {\em Wireless Personal Communications}, vol. 129, pp. 2525--2561,
  2023.

\bibitem{de2023ongoing}
J.~Tian, D.~Hu, et~al.,
\newblock ``Semi-supervised multimodal emotion recognition with consensus
  decision-making and label correction,''
\newblock {\em Proc. MRAC}, pp. 67--73, 2023.

\bibitem{ghosh2022mmer}
S.~Ghosh, U.~Tyagi, et~al.,
\newblock ``Mmer: Multimodal multi-task learning for speech emotion
  recognition,''
\newblock {\em Proc. Interspeech}, pp. 1209--1213, 2023.

\bibitem{liu2022group}
P.~Liu, K.~Li, et~al.,
\newblock ``Group gated fusion on attention-based bidirectional alignment for
  multimodal emotion recognition,''
\newblock {\em Proc. Interspeech}, pp. 379--383, 2020.

\bibitem{fan2022isnet}
W.~Fan, X.~Xu, et~al.,
\newblock ``Isnet: Individual standardization network for speech emotion
  recognition,''
\newblock {\em IEEE/ACM Transactions on Audio, Speech, and Language
  Processing}, vol. 30, pp. 1803--1814, 2022.

\bibitem{chen2022speechformer}
W.~Chen, X.~Xing, et~al.,
\newblock ``Speechformer: A hierarchical efficient framework incorporating the
  characteristics of speech,''
\newblock {\em Proc. Interspeech}, pp. 346--350, 2022.

\bibitem{fan2023mgat}
W.~Fan, X.~Xing, et~al.,
\newblock ``Mgat: Multi-granularity attention based transformers for
  multi-modal emotion recognition,''
\newblock {\em Proc. ICASSP}, pp. 1--5, 2023.

\bibitem{schneider2019wav2vec}
S.~Schneider, A.~Baevski, et~al.,
\newblock ``wav2vec: Unsupervised pre-training for speech recognition,''
\newblock {\em Proc. Interspeech}, pp. 3465--3469, 2019.

\bibitem{hsu2021hubert}
W.-N. Hsu, B.~Bolte, et~al.,
\newblock ``Hubert: Self-supervised speech representation learning by masked
  prediction of hidden units,''
\newblock {\em IEEE/ACM Transactions on Audio, Speech, and Language
  Processing}, vol. 29, pp. 3451--3460, 2021.

\bibitem{chen2022wavlm}
S.~Chen, C.~Wang, et~al.,
\newblock ``Wavlm: Large-scale self-supervised pre-training for full stack
  speech processing,''
\newblock {\em IEEE Journal of Selected Topics in Signal Processing}, vol. 16,
  pp. 1505--1518, 2022.

\bibitem{devlin2018bert}
J.~Devlin, M.-W. Chang, et~al.,
\newblock ``Bert: Pre-training of deep bidirectional transformers for language
  understanding,''
\newblock {\em Proc. NAACL-HLT}, p. 4171–4186, 2019.

\bibitem{he2020deberta}
P.~He, X.~Liu, et~al.,
\newblock ``Deberta: Decoding-enhanced bert with disentangled attention,''
\newblock {\em Proc. ICLR}, 2021.

\bibitem{liu1907roberta}
Y.~Liu, M.~Ott, et~al.,
\newblock ``Roberta: a robustly optimized bert pretraining approach,''
\newblock {\em Proc. ICLR}, 2020.

\bibitem{lin2023robust}
B.~Lin and L.~Wang,
\newblock ``Robust multi-modal speech emotion recognition with asr error
  adaptation,''
\newblock {\em Proc. ICASSP}, pp. 1--5, 2023.

\bibitem{shi2023effectiveness}
X.~Shi, J.~He, et~al.,
\newblock ``On the effectiveness of asr representations in real-world noisy
  speech emotion recognition,''
\newblock {\em arXiv:2311.07093}, 2023.

\bibitem{santoso2022speech}
J.~Santoso, T.~Yamada, et~al.,
\newblock ``Speech emotion recognition based on self-attention weight
  correction for acoustic and text features,''
\newblock {\em IEEE Access}, vol. 10, pp. 115732--115743, 2022.

\bibitem{tripathi2018multi}
Y~Gao, J.~Liu, et~al.,
\newblock ``Domain-adversarial autoencoder with attention based feature level
  fusion for speech emotion recognition,''
\newblock {\em Proc. ICASSP}, pp. 6314--6318, 2021.

\bibitem{choi2018convolutional}
J.~Liu, Z.~Liu, et~al.,
\newblock ``Temporal attention convolutional network for speech emotion
  recognition with latent representation,''
\newblock {\em Proc. Interspeech}, pp. 2337--2341, 2020.

\bibitem{krishna2020multimodal}
D.~Krishna and A.~Patil,
\newblock ``Multimodal emotion recognition using cross-modal attention and 1d
  convolutional neural networks,''
\newblock {\em Proc. Interspeech}, pp. 4243--4247, 2020.

\bibitem{hu2023mir}
Y.~Hu, C.~Chen, et~al.,
\newblock ``Mir-gan: Refining frame-level modality-invariant representations
  with adversarial network for audio-visual speech recognition,''
\newblock {\em Proc. ACL}, pp. 11610--11625, 2023.

\bibitem{hazarika2020misa}
D.~Hazarika, R.~Zimmermann, et~al.,
\newblock ``Misa: Modality-invariant and-specific representations for
  multimodal sentiment analysis,''
\newblock {\em Proc. ACMMM}, pp. 1122--1131, 2020.

\bibitem{yu2021learning}
W.~Yu, H.~Xu, et~al.,
\newblock ``Learning modality-specific representations with self-supervised
  multi-task learning for multimodal sentiment analysis,''
\newblock {\em Proc. AAAI}, pp. 10790--10797, 2021.

\bibitem{yao2022modality}
Y.~Yao and R.~Mihalcea,
\newblock ``Modality-specific learning rates for effective multimodal additive
  late-fusion,''
\newblock {\em Proc. ACL}, pp. 1824--1834, 2022.

\bibitem{yang2022learning}
D.~Yang, H.~Kuang, et~al.,
\newblock ``Learning modality-specific and-agnostic representations for
  asynchronous multimodal language sequences,''
\newblock {\em Proc. ACMMM}, pp. 1708--1717, 2022.

\bibitem{wu2016google}
Y.~Wu, M.~Schuster, et~al.,
\newblock ``Google's neural machine translation system: Bridging the gap
  between human and machine translation,''
\newblock {\em arXiv:1609.08144}, 2016.

\bibitem{DBLP:journals/corr/abs-2208-04641}
J.~He, Z.~Yang, et~al.,
\newblock ``{ED-CEC}: Improving rare word recognition using asr postprocessing
  based on error detection and context-aware error correction,''
\newblock {\em Proc. ASRU}, pp. 1--6, 2023.

\bibitem{vaswani2017attention}
A.~Vaswani, N.~Shazeer, et~al.,
\newblock ``Attention is all you need,''
\newblock {\em Proc. NeurIPS}, vol. 30, pp. 6000--6010, 2017.

\bibitem{tsai2019multimodal}
Y.-H.~H. Tsai, S.~Bai, et~al.,
\newblock ``Multimodal transformer for unaligned multimodal language
  sequences,''
\newblock {\em Proc. ACL}, pp. 6558--6569, 2019.

\bibitem{ba2016layer}
J.~L. Ba, J.~R. Kiros, et~al.,
\newblock ``Layer normalization,''
\newblock {\em arXiv:1607.06450}, 2016.

\bibitem{he2015delving}
K.~He, X.~Zhang, et~al.,
\newblock ``Delving deep into rectifiers: Surpassing human-level performance on
  imagenet classification,''
\newblock {\em Proc. ICCV}, pp. 1026--1034, 2015.

\bibitem{kingma2014adam}
D.~P Kingma and J.~Ba,
\newblock ``Adam: A method for stochastic optimization,''
\newblock {\em Proc. ICLR}, pp. 1--15, 2015.

\bibitem{busso2008iemocap}
C.~Busso, M.~Bulut, et~al.,
\newblock ``{IEMOCAP}: Interactive emotional dyadic motion capture database,''
\newblock {\em Language resources and evaluation}, pp. 335--359, 2008.

\bibitem{radford2022robust}
A.~Radford, J.~W. Kim, et~al.,
\newblock ``Robust speech recognition via large-scale weak supervision,''
\newblock {\em Proc. ICML}, 2023.

\bibitem{leng2021fastcorrect}
Y.~Leng, X.~Tan, et~al.,
\newblock ``Fastcorrect: Fast error correction with edit alignment for
  automatic speech recognition,''
\newblock {\em Proc. NeurIPS}, pp. 21708--21719, 2021.

\end{thebibliography}

\end{document}